# My Life in Artificial Intelligence

*People, anecdotes, and some lessons learnt*

By Kees van Deemter

**Abstract**: In this very personal "workography", I relate my 40-year experiences as a researcher and educator in and around Artificial Intelligence (AI), more specifically Natural Language Processing. I describe how curiosity, and the circumstances of the day, led me to work in both industry and academia, and in various countries, including The Netherlands (Amsterdam, Eindhoven, and Utrecht), the USA (Stanford), England (Brighton), Scotland (Aberdeen), and China (Beijing and Harbin). People and anecdotes play a large role in my story; the history of AI forms its backdrop. I focus on things that might be of interest to (even) younger colleagues, given the choices they face in their own work and life at a time when AI is finally emerging from the shadows.

\

## 1. Aim and structure of this workography.

I've had the privilege of working as a researcher and educator in Artificial Intelligence (AI) and related subjects for about 40 years. I'm nothing out of the ordinary; some of my peers have achieved far more than I. Still, I'd like to jot down some of my experiences, because they are likely to differ from those of most people, and because few areas of work have changed so rapidly over these years as AI. Perhaps my "(auto)workography" can help younger colleagues to avoid some of the mistakes I made. Others might simply enjoy reading some stories about a bygone era, which feature some remarkable people: the good, the bad, and the ugly.

The resulting document is quite a personal one, which should not be read as a mini-treatise on AI or any part of it. I will occasionally try to convey the flavour of the work that my colleagues and I have been doing, but always only in passing. To put things in context, two short sections, at the beginning and at the end of this text, describe the history of AI from the 1950s onward, and the position in which AI research finds itself at the moment. In case you want to find out more, I'll refer to the scholarly literature now and then, but most people will want to ignore these pointers.

I start out with a brief introduction to AI and Natural Language Processing (Section 2). After that, I'll give a chronological account of my experiences in Amsterdam (Section 3), Eindhoven (Section 4), Amsterdam again and Stanford (Section 5), Eindhoven again and Brighton (Section 6), Aberdeen (Section 7), in China (Section 8), and finally in Utrecht (Section 9). After a brief discussion of some loose ends (Section 10), I conclude by revisiting the changes that have taken place in AI and NLP over the last 40 years, and expressing some thoughts concerning the way forward (Section 11).

## 2. Introduction: Artificial Intelligence and NLP

Artificial Intelligence (AI) is older than many people realize. For instance, as early as 1950, Alan Turing, one of the fathers of computing science, published some influential ideas on what it would take for computers to "think".[1] The AI research programme, as a systematic pursuit, was launched at the famous Dartmouth Conference in 1956. When proposing the conference, the initiators wrote that, in this fledgling new area of research, *"An attempt will be made to find how to make machines use language, form abstractions and concepts, solve problems now reserved for humans, and improve themselves."*[2] Since then, AI has never gone away. AI can be engineering or science. In the former case, AI is pursued for practical reasons, to build useful tools. In the latter, the aim of AI is to create working models of human behaviour, in order to expound and test scientific hypotheses about what and who we are as human beings. As you can imagine, adherents of each of these two brands of AI sometimes consider their own to be the only “real” AI, a phenomenon that will be one of the emerging themes of this workography.

[1] Turing (1950).
[2] For a reprint of the conference proposal, see McCarthy et al. (1955).

Natural Language Processing (NLP) has always been at the core of AI, as was the idea of letting computers learn from, and reason about, data. Researchers in NLP construct computer programs that can *generate* language, that is, they can write and/or speak. This is called Natural Language Generation (NLG).[3] NLP programs can *interpret* language as well, which is called Natural Language Understanding. Key applications of NLP include Machine Translation, Question Answering (i.e., letting the computer answer questions posed by a person), Automatic Summarization, and so on. AI comprises many other activities, but my own work has been in NLP, focussing mostly on NLG. I can’t think of any ”deep” reason for my focus on NLG instead of other areas of NLP; it’s simply the topic that came knocking on my door.

Over time, the ambitious research program of AI has gone through ups and downs. Famous *ups* include the invention of logic programming languages such as PROLOG, which powered the Japanese Fifth-Generation initiative of the 1980s.[4] Notorious *downs* include a long “AI winter”, at the end of the 20$^{th}$ century, when funding for AI dried up. Among other things, the AI winter was caused by the inability of AI programs to see patterns, and to bring common sense and background knowledge to bear on the interpretation and generation of text, as was argued in a rather devastating report for the US government.[5] For a long time, this report, and the observations that were made in it, had a chilling effect on the development of AI.

Sometime around 1990, AI researchers started to move away from rule-based algorithms, focussing more and more on letting computers learn from data “by themselves”, using a variety of statistical methods. Ideas along these lines had been around for decades[6], but now they were catching on more broadly. Aided by rapidly growing computing power, statistical methods have recently culminated in computer programs such as ChatGPT,[7] which solve NLP problems by learning from massive amounts of textual and other data. As if by magic, these methods give AI programs a much better handle on the limitations that had plagued older AI programs, subtly leveraging lots of information that is somehow buried in all that data from which these programs learn. Despite the errors that they frequently still commit, these programs are widely seen as successful, including – and perhaps especially – by non-expert users.

ChatGPT-like systems produce textual and other output, conditioned on a user-provided prompt, which can be a question, a command, or any amount of other text. Where this involves generating textual output, these systems can be seen as performing Natural Language Generation (NLG), among some other tasks. Since they can produce images and programming code and sounds as well as text, they are now frequently referred to by the more wide-ranging term "Generative AI".

---

[3] See Gatt and Krahmer (2018) for a survey of NLG. Reiter (2024) for a brand-new monograph on NLG.
[4] See Feigenbaum and McCorduck (1983) for an overview of the Sixth-Generation project.
[5] Bar Hillel (1960).
[6] See, for example, Weaver (1953), Rosenblatt (1958).
[7] See Roumeliotis et al. (2023) for discussion.

I entered the field of AI in the 1980s. Having become an Emeritus Professor in 2024, I'm preparing to leave the field just when the AI winter appears to be over. This is an interesting time to look back. Before briefly assessing the state of AI *anno* 2025 (Section 11), I will narrate my own journey, using as my anchors the places where I plied my trade. I'll tell a few anecdotes to keep you awake, and I'll mention a few people who have been important to me, leaving out many more who have been important to me as well.

## 3. Studying at the University of Amsterdam (1979-1983)

I did not start out my academic journey in AI. In fact, as an 18-year old, I started out studying philosophy, at Leiden University in The Netherlands. But after about a year, I knew that speculative philosophy was not my thing; I was more interested in human reasoning, and hence in both mathematical logic and human language. These topics are addressed in most philosophy courses, including in Leiden, but I soon realized that a multidisciplinary set of people at the University of Amsterdam, at what is now the Institute for Logic, Language and Computation,[8] were at the forefront of this work, which is associated with the term *formal semantics (of natural language)*.[9] I decided to transfer to Amsterdam. Although this was easy in administrative terms, intellectually it was challenging, because I had missed out on some of the necessary maths.

After a year or so, I got the hang of things, helped along by the classes taught by Frank Veltman and others. Frank used these classes to let us think through in detail how logical expressions can formalise abstract ideas. I remember one course in particular, which concerned different assumptions about the structure of time – Is time ordered continuously? Does it have a beginning? An end? – and how these assumptions can be formalised. Far from merely presenting a prefabricated set of ideas, he was thinking on his feet, forcing us, his students, to think as well, and to experiment with formal mechanisms. I think I've never seen anyone teach as well.

Talking about people who taught me, I also want to mention my math teacher in high school, Mr. van Wijk, who insisted that the best way to understand a maths text is to "do it yourself". When you've read a theorem in a book, don't read the proof immediately; it's more instructive to try proving the theorem yourself first. If you're able to prove it, this will give you satisfaction; if you're not, your attempt will put you in a better position to appreciate the proof. Either way, you'll understand and remember the proof more easily than if you'd swallowed it passively.[10] I don't remember how much I heeded this advice at the time – I only remember that I once accidentally crashed my bicycle into Mr. van Wijk's Citroën 2CV car, scarring the hood of his car as much as my own chin – but it's been useful to me ever since I got into maths more deeply as a university student. Gradually, I became a part of the furniture of the department in Amsterdam, and the people working there ended up influencing me considerably.[11] Their work combined linguistic, mathematical, and engineering strands into a remarkably coherent enterprise, inspired by the works of Richard Montague and others,[12] which was gaining interest in academic circles around the world. It was an exciting environment in which to be a student. After a difficult start, my move to Amsterdam was finally vindicated.

---

[8] https://www.illc.uva.nl .

[9] Formal semanticists study the meaning of sentences and texts by linking them systematically with formulas in logic, or with ideas in other areas of maths, such as Game Theory.

A key moment came when I was hired as a student assistant. For a start, it allowed me to pay off some considerable debts that I had incurred, and that had worried me greatly; finally, I felt some semblance of order returning to my life! Just as important, the group in Amsterdam often had lunch together, carrying on lively discussions, so I could witness first hand who these people really were, and how they thought. The way the group conducted their business, with curiosity, enthusiasm, and camaraderie, has always remained something of a model for me, even when my research drifted off in a different direction. Comparing these experiences with our own students in Utrecht's Computing Science department *anno* 2025, where everything happens at such a scale that very few students get to really know their lecturers, I realize how fortunate I was as a student.

This does not mean everything was plain sailing. On the first day of my job as a student assistant, the Amsterdam Colloquium was starting, a yearly event that attracted formal semanticists from all over the world, including such celebrities as Hans Kamp, Barbara Partee, and David Lewis. Early one morning, Frank Veltman, who was one of the organisers, walked up to me and said, "Ah, there you are! We have a bit of an emergency on our hands! Here are some handout papers for the people in the auditorium, would you mind photocopying these to make 100 booklets? Please?" I ran towards the building where the photocopying machines were, keen to make myself useful. After about 15 minutes, Fred Landman entered, who had once held the same assistantship and had now gone on to bigger and better things. "Photocopying? Is that what you think this job is about?" He smiled mischievously. "Can I give you some advice? Do it badly!" I thought it best not to follow his advice, so I finished the job as well as I could, I rushed back to the auditorium and had the handouts distributed. Job done! Except that, despite my best efforts, I must have done something wrong… For although most of the booklets had the correct number of pages, the last few words and symbols of every line on every page were missing. Hardly a sentence could be understood, so the people in the room discovered, one by one, that their handouts were completely useless. I was mortified!

[10] An interesting situation arises if your proof differs from the one in the book, in which case you should compare the two proofs. Did you make a mistake? Or did you find an alternative way to prove the theorem?
[11] I'm thinking of Frank Veltman, Jeroen Groenendijk, Martin Stokhof, Theo Janssen, Roel de Vrijer, and Renate Bartsch. Johan van Benthem, still based in Groningen at the time, was to join them a few years later, in 1986.
[12] Montague's famous research papers were posthumously collected in Thomason (1974). Paul Piwek recently pointed me to https://www.richardmontague.com as a source of information about Montague's life.

No-one told me off, and no-one asked me to make photocopies ever again.

## 4. First full-time job (1984-1997): Researcher at the Institute for Perception Research (Philips/IPO) in Eindhoven

A few years later, at a later instalment of the Amsterdam Colloquium, the person next to me – I think it was Remko Scha – was scribbling in a notebook that had the logo of Philips Electronics on it. Surprised to see someone from industry attending a conference on formal semantics, I asked why he was here. I learnt that people at Philips were applying formal semantics to the construction of Question-Answering systems;[13] the idea was to map any question that a user might ask to an expression in mathematical logic. Since I was starting to think about jobs, I decided to write a letter to the director of Philips's famous Eindhoven NatLab.[14], to ask whether they might have a job for me, although no vacancy had been advertised. The year was 1983, and the world was in the throes of a severe economic downturn; the jobless rate in The Netherlands was above 10%. My letter was obviously a long shot. Miraculously though, I was invited for a series of job interviews in Eindhoven, which I battled through in a jacket borrowed from a friend. At the end of a long day, I was offered a job as a researcher at the Institute for Perception Research (IPO), in which the NatLab collaborated with Eindhoven University of Technology. I was lucky they offered me the job, the more so because my relative lack of training in Computing Science – I had completed only one course in that area, in which I had learnt to program in Pascal – meant I'd have to go on a lengthy course first.

Having been offered the job, I asked Martin Stokhof in Amsterdam, whom I was helping to teach a class on Montague Grammar (having graduated from my stint as an unreliable photocopying agent) whether he thought I should accept. Would it be worthwhile working in industry? Martin responded with great foresight. He argued that, as analyses in formal semantics were becoming more and more complex, they would benefit from being implemented in computer programs, being tested, then modified where necessary; researchers in companies like Philips were well placed to do all that. After thinking things over for a while, I accepted the offer. I knew that it might transform me from a theoretician into something of an engineer, but this was a metamorphosis I could live with – or so I thought.

I had good years at IPO, helped by great colleagues, such as John de Vet, who was able to grasp vast amounts of complex Pascal computer code, Robbert-Jan Beun, a great sparring partner in discussions on any topic, and Paul Piwek[15], who always thinks things through one step deeper than others. John's and my research – or maybe "development" would be a better way to describe the work we were doing at the time – focussed in part on circumstances in which a user's question can be interpreted in more than one way, or when there's something else wrong with the question. If the user asks, "What was the title of Allen Turing's 1950 journal paper?", a really good answer could go, "Do you mean Alan Turing? (Note the different spelling of his name.) The title of his 1950 paper in the journal *Mind* is *Computing Machinery and Intelligence*." The answers produced by our system, which was called SPICOS, were less sophisticated, but they were more informative, and less misleading, than the straightforward answer *"The title is Computing Machinery and Intelligence"* would have been.

[13] Philips' logic-based approach to Question Answering is described briefly in Bronnenberg et al. (1980) and more elaborately in Scha (1983).
[14] Some of the NatLab's main claims to fame lie in low-temperature physics. For a history of the Philips NatLab, see De Vries (2005).

I recently pondered Martin Stokhof's words again, about the choice between academia and industry. His remarks about implementation and evaluation of theories were spot on; this is now obvious to anyone observing recent developments, where the evaluation of Generative AI is so crucial.[16] However, there is another side to this coin: universities tend to be stabler than companies. Companies and research institutes fall over or change course. Universities seldom do, because they're engaged in teaching as well as research, and teaching tends to be a more stable source of income.

I experienced the downside of industry in the years following my PhD (Sections 5 and 6), when Philips found itself in financial difficulties. The then president, Jan Timmer, felt that the best way to save a tech company is … to do more and better tech. Under his guidance, several potential solutions for the woes of the company were attempted. We as researchers tended to appreciate Timmer's combative approach; on the other hand, each new solution forced us to change course. In my case, this meant spending a year fiddling with a fancy piece of audio equipment, trying to improve its user interface. This was not something I was equipped to do very well, no doubt to the occasional frustration of the leader of the project, Jack Gerrissen. Our little team ploughed on for about a year, until Philips decided that the equipment we were working on was never going to be commercially viable, regardless of what improvements we might be able to propose. We were asked to drop everything and start a new line of work.

Rapid changes of direction may sometimes be necessary to keep a company afloat, but they're not ideal for a young researcher who is trying to find a research *niche* of their own. Universities are a much better basis from which to do that. Moreover, in academia, you're essentially your own boss; your superiors are, first and foremost, coordinators and advisors. (If they don't know this, you should tell them :-) In industry, ultimately, the company director is your boss; and not everyone is cut out to be an obedient employee.

---

[15] See e.g. Piwek et al. (2008). Paul joined us at the ITRI in Brighton in 1998 (see Section 6).
[16] See, for example, Mao et al. (2024) for recent tests of the abilities of LLMs.

At the time of writing, and many changes of direction later, Philips has weathered many financial storms but its role in scientific research is much reduced from what it once was.

Within Philips, IPO took up a special position, because of its collaboration with Eindhoven University; in many ways, the IPO felt like a university department. The more research labs I've seen, the more I see how remarkable the IPO was. It was not only a place where theoreticians and technicians, from both industry and academia, were conducting high-quality collaborative research, in a fancy new building especially designed for delicate experimental work. It was also a place that was as cleverly organized as any I've seen. In order to create synergies within an intellectually diverse workforce, some highly effective mechanisms had been put in place: In the main hallway stood a panel that showed, at a glance, who was in and who was out of the building; an intercom was used for announcing research seminars and other communal events; every morning at a fixed time, the intercom invited everyone to take part in 15 minutes of joint coffee in the large canteen. There was an unwritten rule that, if your diary allowed it, you should attend; the same was true for departmental seminars. All these things contributed to an atmosphere that maximized opportunities for learning about each other's work, and for socializing as well. There was even an IPO band that performed at parties! I still meet some of my former IPO colleagues regularly.[17]

Over the years, I've found that the humble learn quickly. With hindsight, at this stage of my career, I was not humble enough: I believed, quite absolutely, in the skills I had learnt in Amsterdam, and it took me a long time to absorb the skills of my new colleagues. For example, IPO was full of people who excelled in experimental design and hypothesis testing. I should have learnt from them, but I only started making an effort after a number of years. It's a gap in my knowledge that I've only much later been able to plug.

One of the people I enjoyed talking to was Sieb Nooteboom, who led IPO's Speech group at the time. I remember how, more or less by accident, he attended a research talk I gave at Tilburg University one day, when I was visiting the group led by Harry Bunt. After the talk, Sieb walked up to me and said, "Nice talk, Kees, but you crammed in far too much. People are stupid. If you give a research talk, you're going to get across at most one message. At most! If you try more than one, you risk conveying nothing at all." I think this was excellent advice. Sieb was no shrinking violet though. I remember him saying, "Someone who asks my advice should always follow it". I thought this was not such good advice, for instance because it suggests that once he had given you advice on a given question, you should never ask anyone else's.

Another source of insight was the director of the IPO, Herman Bouma, who led the Institute in exemplary fashion. Sometimes he found opportunities to teach us some lessons, for instance in his meetings with IPO's workers' union, which I chaired for a few years. Here is one such lesson (reproduced as faithfully as memory allows): When you're young, you should work to improve in your areas of weakness; when you're older, you should do the things you're good at. – Maybe now, at 68, is a good time for me to finally take the second part of this lesson to heart.

[17] As will become clear in Section 6, I meet one of them *very* regularly.

## 5. PhD student in Amsterdam; Postdoc at Stanford (1987-1993).

While being employed by Philips in Eindhoven, I was able to do a PhD in Amsterdam, followed by a postdoc at Stanford, California. I'm still grateful for these opportunities, so let me explain how they came about.

One day in Eindhoven, the IPO was visited by someone I had met a couple of times when working on (what would now be called) my Master Thesis. Johan van Benthem, an authority in modal logic, whose academic lineage descended from Bertrand Russell and Ludwig Wittgenstein, had moved from Groningen to Amsterdam. His interests went beyond pure maths, including the various ways in which mathematical logic can be employed to analyse language. At some point, he visited the IPO. Around 5 o'clock in the afternoon, he and the Institute's director, Herman Bouma knocked on my door. After a brief chat, Johan asked me whether I wanted to do a PhD with him, while continuing to work for Philips; clearly, Herman Bouma was in on the act. I seldom take decisions on the spur of the moment but on this occasion, I think I consented immediately.

Doing a PhD while working for a company creates a situation in which your interests and those of the company are not very well aligned. For, at the end of the day, the company wants to sell products and keep some of their knowledge a secret, whereas you, as a researcher, care more about answering research questions, and getting your answers published. Luckily, I had the best thesis supervisor I could have wished for. One useful piece of advice from him was that you should not merely focus on the topic you happen to work on at the moment; you should always be learning about new ideas and techniques, in systematic fashion, for example by working through a difficult book.

Johan and I only met every 3 or 4 months, usually in Amsterdam, relying mostly on postal mail for additional communication. Although this frequency is sparse by today's standards, he had a knack for making it work. Before we met, I'd send him some notes, for example a rough draft of a thesis chapter. A few days later, I'd take a train to Amsterdam, where we'd discuss things for an hour or so, after which I'd return to Eindhoven. Then, a few days after our meeting, a letter would fall onto my doormat, containing some further thoughts regarding my chapter, presented in Johan's famously meticulous handwriting, in which each letter was drawn individually, resembling printed text. (I still have a few of these letters in my drawer, waiting for their money value to peak.) Johan's standard procedure of discussing my work in two steps, first oral then written, worked well for me; the first stage is for getting to understand what your student has in mind, the second for putting them straight. (I occasionally wonder: why do we do things so differently these days, meeting with our PhD students every week?)

In my work with Johan, I studied some theoretical aspects of Question Answering; for instance, what strategies are open to a Question Answering system when a user asks a question that can be interpreted in different ways? How does logical deduction behave in such situations?[18] My thesis, as a whole, was much more typical of the work that is done at a university than of the work that is done at the research department of a company. I learnt a lot from it, though it's doubtful that Philips benefitted much.

Whereas my research may have proceeded a bit more slowly than would have been the case in a purely academic setting, the challenges inherent in my position – somewhere in between academia and industry – allowed me to learn quite a bit about life. I often had to justify my work to engineers and psychologists, and my position forced me to improve my planning skills. Dwight Eisenhower, former WW-II general and 34th President of the US, is credited with saying, "In preparing for battle I have always found that plans are useless, but planning is indispensable". I don't know much about battle, but Eisenhower's saying is very apt for scientific research. For, on the one hand, research is often open ended, because its future direction is influenced by what you find now; ideas from other people can likewise influence your direction, so all plans are a hostage to fortune. Yet, without a plan, you're not going to achieve a coherent body of work. It's therefore important to spend time making plans, while always being prepared to modify them.

After my PhD, while still working at IPO, I started looking for opportunities to extend my experience. I'd already spent a very useful few weeks at the 1987 LSA Linguistics Institute, which was held at Stanford – I particularly remember a course on Irene Heim's File Change semantics by the famous Barbara Partee, which had given me some crucial inspiration for my own PhD work – and now Johan suggested that a longer period at Stanford, where he himself was working one semester every year, might be useful.

This was the start of an interesting tussle. People higher up in the Philips food chain were opposed to the plan, because they felt – understandable from their perspective – that it was time for me to start earning some money for them. Things were dragging on. At one point, I was so sick of the delays that I almost literally couldn't get out of bed on a day when there was an important IPO-wide gathering. Noticing my return to work the next day, Herman Bouma asked, "Were you ill yesterday?" I responded, "I had serious complaints!" *("Ik had ernstige klachten!")* I think he understood my feelings very well.

A Daoist thinker in ancient China once argued that the lowly ant is more powerful than the mighty lion. For example, a wooden gate might successfully lock in a lion but not an ant, because the latter will crawl underneath the woodwork unimpeded. Similarly, in a multinational company, the lowly employee can sometimes be more powerful than the mighty director. Suppose you, the employee, have a plan that's important to you personally (like my plan of visiting Stanford was for me). A company director may reject your plan but 10 minutes later, he has other things on his mind, and the next day he's forgotten about you entirely. But since the matter is important to you, *you* do not forget. Sooner or later you'll find a way to get what you want!

---

[18] I think this question has not yet found a totally satisfactory answer in mathematical logic even now, but modern NLP systems make an interesting stab at some closely related problems when they perform Natural Language Inference (NLI). How successful they are in this specific regard is a question worth investigating.

I do not know what happened behind the scenes but, after a brief standoff, a compromise was reached: I was allowed to visit Stanford for a year, but I wouldn't get any help from Philips. In practice, this meant I had to apply for government funding. I decided to try for a so-called NWO Talent grant, proposing a project on reasoning with ambiguous logical formulas. To my delight, my proposal was funded. To complete my victory, Stanley Peters, an eminent formal semanticist, told me I was welcome to work on my Talent project as a "visiting scholar" (essentially a postdoc) in his group at Stanford.

When I told my Philips bosses about these developments, while trying to refrain from gloating, they gave in. They suggested that I discuss some practicalities with an administrative department somewhere in a remote corner of the city. I remember stepping off my bicycle in front of a wind-swept building where there were no other bicycles to be seen – a sure sign I had arrived in another part of town. The elevator took ages to reach the top of the building. I knocked on an arbitrary door and I entered an office where I met a friendly man, who took the time to hear me out. Somewhere along the way, he must have thought, "This poor guy needs help!", because after this meeting, I was given all the support one can imagine, financially and otherwise, without ever having asked for it.

Stanford proved to be a stimulating environment. Its campus is situated in a leafy area not far from the Pacific, between rolling hills, and only about an hour's ride from bustling San Francisco. Most important of all, Stanford was and is full of people who live and breathe research. During my time there, I was able to elaborate on my PhD work, for example by studying implications for speech science, and by studying the challenges that arise when mathematical logic is employed to formalise the meaning of sentences that are not very clear, for example because they contain vague or ambiguous words. This is not a task mathematical logic was originally invented to perform, but it turned out that, if you force it to, it can do the job. I benefitted greatly from discussing these and other matters with the locals. Henriette de Swart and Cleo Condoravdi, for example, disabused me gently but thoroughly of the, evidently preposterous, notion that I knew all about linguistics that was worth knowing, while Ed Zalta – who was kind enough to share his "Metaphysics Research Lab" with me – taught me as much about music and mountain hiking as about philosophical logic. At the same time, friends and family took turns visiting me, enjoying with me all that California had to offer. In short, I had a ball!

One lasting impression from my year at Stanford is how the research community there consisted not only of the famous people who were household names already,[19] but also of a sizable set of people who were so drawn to this intellectual magnet that they spent years of their lives taking part in research seminars and other on-campus events. Many of them were to become established figures. It was clear that theirs was not just a prudent gamble: they were at Stanford because, as long as they could afford it, to work there seemed to them to be the best way to spend their lives. I do not disagree with them.

Most people at Stanford worked as if their lives depended on it. A real or apparent exception to this rule was Stanley Peters. One of my self-imposed tasks was to organise the weekly sessions of a reading group on formal semantics, which took place in a kind of living room in Ventura Hall, which looked out over a nice garden. The idea was for all participants to read a research paper beforehand, then to discuss it together during the meeting. Stanley didn't always come well prepared, but he made up for this by subjecting the other participants to some pointed questions about the paper, which we'd do our best to answer. Now and then, one of us would walk over to the whiteboard to explain one of the finer points and, as the meeting progressed, our understanding of the paper, including its weaknesses, grew. In many ways, Stanley's questions were a master class in dissecting a piece of research. It was particularly instructive to see how he was never afraid of appearing stupid or ignorant; at the end of the meeting, he often understood the paper better than anyone else in the room.

A side-effect of my stay at Stanford was to see how famous the people who had taught me in Amsterdam were there. I remember this about Martin Stokhof and Jeroen Groenendijk in particular. I'd always had a very high regard for them, but I never realized how famous they were; I needed the people at Stanford to tell me that.

## 6. Second stint at Philips (1993-1996); Researcher at the University of Brighton (1997-2004)

Still at Stanford, thinking about my imminent return to IPO, I was faced with a difficult decision. After returning to The Netherlands, should I return to the group of Don Bouwhuis, a fine experimental psychologist who led a group on Cognition and Communication? I'd been a member of Don's group before, and I knew that he gave the people in his group a lot of freedom. He had a great sense of humour that made him a pleasure to work with. In many ways, returning to his group was the easy option.

A more adventurous alternative was for me to join a new group that had recently joined IPO, and that focussed on NLP. The group, led by Jan Landsbergen, had previously worked on Machine Translation but were now asked by Philips to make a fresh start working on other NLP problems. Joining them now would be risky, after all Philips had effectively declared their previous work a failure, at least from a commercial point of view. Pondering the situation from faraway Stanford, I admired the work they had done.[20] I felt that, when NLP researchers of the calibre of Jan Landsbergen and Jan Odijk join your department, it would be foolish not to join them. I decided that I had to do it, regardless of the risks. With hindsight, it was the right decision, but I could have done a better job explaining myself to my colleagues. Instead of sending them my thoughts in email from Stanford, I should have realized that difficult messages should not be conveyed by email, but preferably in person, because the immediacy and greater bandwith of a live encounter offer all sides a better chance to avoid misunderstandings.

[19] Some of these "household names" were John McCarthy (author of McCarthy et al. 1955), Ivan Sag, Laurie Karttunen, Annie Zaenen, Terry Winograd, and Pat Suppes. Among the PhD students during my time at Stanford where Chris Manning and Hinrich Schütze, both of whom went on to become highly influential in NLP.

Jan Landsbergen, the head of IPO's new NLP group, had stood at the cradle of some of the most elegantly conceived NLP projects that I have seen. The Rosetta project, for example, put Montague's framework on its head in a way that very few people could have thought up, maintaining the algebraic structure of the framework but abolishing everything that was not necessary for performing Machine Translation. A striking quality of Jan, as a group leader, is how he was always honest, for example when presenting the work of our group, regardless of who was listening. Rather than "spinning" our work to make it look good, he'd dwell on things that were not going so well, or about which we were still unsure. In later life, I may have tried to emulate his approach, since someone who knows us both sometimes accuses me of "doing a Landsbergen" when she feels I'm engaged in an act of professional self harm.

A notable moment came on the day of my return to Eindhoven. I entered the room of an American colleague, Teddy McCalley, to tell her about my Californian adventure. Reassuringly, her room was filled with smoke, as it had always been. We had much to talk about, but I was visually distracted by the presence of another person in the room, a young PhD student named Judith Masthoff, who had joined the IPO, and Teddy's office, during my long absence. Soon Judith and I were talking in the stairway, leaning against the railing on opposite sides of the stairs. We married in 2001, and today we're still together. A few more things could be said about the matter, as you can imagine, but I gather that a workography is not the best place for that.

After my return to the IPO, as a member of the Jan Landsbergen's NLP group, I worked on an NLG system called Dial Your Disc, that helped users browse music collections. Music was still one of Philips' core activities, and the company wanted to build systems that help users to put together their own music CDs. At each point during a user's interaction with Dial Your Disc, the system would convert information in a database into spoken English to explain to the user what music they were about to hear, and how this music related to previous things they'd heard on the system. For instance, the system might say, "You're now going to listen to Mozart's fourteenth piano sonata. Unlike the sonata that you've heard a few minutes ago, the fourteenth sonata is in C minor …". My own, relatively modest, contribution to this work focussed on deciding what parts of each sentence should carry a pitch accent, combining the work in formal semantics that I'd done at Stanford with work on NLG and speech synthesis that had been done by a two separate groups of people at IPO. The three pieces of work dovetailed remarkably well. For me, the project was my first introduction to NLG – Natural Language Generation, which is the computational generation of text. I didn't know it at the time, but NLG would turn out to be the main focus of my research from that time onwards.

[20] Rosetta (1994) offers a broad overview of the entire Rosetta Machine Translation project.

Unfortunately, the Dial Your Disc project didn't last very long. Philips had supported NLP research with great generosity from about 1974 onwards, but they had never reaped much financial reward. With impeccable timing, the company abandoned ship when NLP began to be commercially successful. The divorce started messy, but in the end my colleagues and I were treated in gentlemanly fashion. I knew that academia had always been a more natural home for me than industry, so I applied for academic jobs in the main. Since, like most of my Philips colleagues at the time, I hadn't published very much, I made a slow start until I found out two things. First, people who do not know you in person are unlikely to hire you. Second, if you're in need of a job, you need to shout about it; a mere whisper is not enough. Armed with these lessons, I found a handful of job opportunities, in four different countries. How to make a choice?

One job offer took me a while to sort out. At Austin, Texas there was, and still is, a company called Cycorp. The company had grown out of an academic research project called CYC which aimed, ultimately, to formalize all encyclopedic knowledge using logical axioms. This is obviously an gigantic task, but the founder of both CYC and Cycorp, Douglas Lenat, believed that it was feasible and he called it "real AI" because, basically – Don't you know? – nothing else in AI was real. While I was hesitant about this claim, I saw the attraction of working with their ideas.

I visited Austin with Judith (who was applying for another job there), but after a few days, we left somewhat baffled. I was offered a well-paid job, but Doug Lenat proved to be quite a character. Sat next to me in his extremely sporty car, driving through a landscape that, even in May, felt tropical to me, Doug confided that he didn't believe in reading books and articles, because most of what was written about AI was, in his modest opinion, total garbage. With only one or two exceptions – he left it unclear whether I'd be one of them – he expected his people to simply put in the hours coding ontologies in the logical formalism that the company had developed; if they did that well, he was happy. Having returned to Eindhoven, I decided that this job was not for me after all. Playing with the CYC mechanism without the freedom to modify it would be interesting for a year maybe, but I feared that it would become mere drudgery after that. I was also disappointed that the company allowed me only about two weeks of Annual Leave, which is not a lot if you've got friends and family overseas. I was used to working hard, and with great enthusiasm, but I could still remember there's more to life than only work.

One of the other jobs on my list was at the Information Technology Research Institute (ITRI) at the University of Brighton, on the English south coast. On paper, it was probably the least prestigious of the five jobs because, as a former Polytechnic, Brighton was more known for solid vocational training than for exciting research. For this reason, some British academics warned me off. Asking around in the academically more egalitarian Netherlands, the advice was more positive. The director of the Institute was Donia Scott, whom I had met a couple of times when she was still working for Philips in England. She enjoyed an excellent reputation as a researcher and a manager. The people she'd surrounded herself with were very smart, and very driven. Their research focussed, among many other things, on a new type of computer interface that helps human users enter complex information by letting an NLG system express this information in ordinary language, at the same time that you're entering it; the name of this clever knowledge editing method, which was being invented at ITRI, was What You See Is What You Meant (WYSIWYM).[21] The Dial Your Disc project at Philips had gotten me interested in NLG, so this job could be a nice fit.

The trip to my job interview came hard on the heels of some conference deadlines. I was tired and I'd been too busy to book a hotel. Right next to Brighton railway station, the people in a less than spotless office pointed me to a Bed & Breakfast somewhere on the windswept seafront, where there seemed to be a vacancy. I had a good night's sleep there and I took a taxi to the University after breakfast. It wasn't far, but I had underestimated the traffic, which came to a standstill several times. Nervous, I arrived about 15 minutes late, but the locals, who knew about the traffic, seemed to forgive me and soon we were engaged in animated conversation. In the end, I was offered the job, and I decided to accept. The pay was below what I'd become used to, and the position was not permanent; also, there was Judith's job situation to be sorted out. However, we both had a feeling that everything would turn out alright.

Given that our lifestyle did not require huge amounts of money – I know it's different if you've got twelve children – I stand by my job-selection attitude at the time. After all, your opportunities for conducting interesting research should count for more than your annual paycheck. (And, if your research is good, the pay is likely to follow.) After about a year and a half, Donia ensured that my position was made permanent. I'm sure that a certain amount of luck was involved, because her move might have been financially impossible had the ITRI been less successful getting grants during that period.

Initially I had worked on incorporating images into the WYSIWYM paradigm but, in the absence of suitable techniques for computationally generating pictures and diagrams (so that all we could do is work with pre-existing images), I found it hard to make a success of this. Therefore, I pivotted towards a different research topic: the design of computer algorithms that mimic what people say when they *refer* to objects, that is, when they verbally identify an object for a hearer, to ensure the hearer knows what they're talking about.[22] Reference is a key mechanism in almost all communication; we use it whenever we speak or write, often many times per sentence. As a research topic within the broader area of NLG, it combined substantial theoretical interest with obvious relevance for practical applications.

[21] See e.g., Power et al. (1998) for a statement of the principles of WYSIWYM-based knowledge editing. The name WYSIWYM alludes to the more widely known WYSIWYG (What You See Is What You Get) text editing paradigm.

As a side effect, my work on reference helped me understand my own strengths and weaknesses better: I do not always enjoy building large systems, with all the cumbersome details that this involves. (I blush while writing this down!) What I do like, however, is trying to see the big picture. What I liked in particular is to use some simple maths to shed light on a piece of research, for example by finding a structural flaw in it. In this way, my ITRI colleague Rodger Kibble and I detected a flaw in the manner in which a prominent strand of work in NLP operationalised the notion of co-reference (i.e., where several expressions refer to one and the same entity).[23] Explaining and analysing this flaw, and discussing remedies against it, turned out to be a rewarding task, which came with a pleasant "David defeats Goliath" feeling when people finally agreed with us.[24]

On the whole, academics make superb colleagues. It would be hard to compose a group of people more interesting than the ones that frequently had lunch around ITRI's minimalistic kitchen table, in the outskirts of Brighton, between the foothills of the South Downs. One colleague who stood out in this respect was Richard Power (see e.g., Power (1979) for his classic work on collaborating AI agents), who was able to use his academic brain to shed light on just about any topic in daily life in a way that was enlightening, original, and often humourous. Depending on who else was around, Roger Evans, Nadjet Bouayad Agha, or Daniel Paiva might join in. It's conversations like the ones we used to have around that kitchen table that I will miss when I'm retired.

Another notable ITRI colleague was Adam Kilgarriff. Adam's specialities were corpus linguistics and lexicography, and his work on corpora had made him an early sceptic about rule-based methods in NLP (see e.g., Kilgarriff 1997). This made him a bit of an outlier within the Institute, in which most others worked on NLG, using rule-based methods. At some point, when I got a nice job promotion, he confided that he found it difficult to be happy for me, because he himself was not getting the recognition he felt he deserved. The years passed, however, and not long before his untimely death in 2015, when I was working in Aberdeen, I discovered that interest in his work had mushroomed! His work had acquired a reknown far greater than that of any of his former ITRI colleagues, for example in connection with the Word Sketch Engine, which helps users search corpora for all occurrences of a given word in which that word is used in a specific syntactic position (Kilgarriff 2004). It's a wonderful tool[25], which I recommend heartily.

[22] See Van Deemter (2016) for a monograph on both computational and experimental studies of the act of referring, discussing these two angles as different parts of one and the same enterprise.
[23] See Van Deemter and Kibble (2000). Coreference is important for many practical applications, such as Information Extraction, for example.
[24] A similar attempt at "flaw finding" is Van Deemter (2024), which analyses some recent proposals for classifying the errors committed by Deep Learning-based NLG systems.

Looking back, Adam had simply been far quicker than the rest of us in recognizing the virtues of the new, statistical methods in AI. When I emailed him to congratulate him on his successes he said, with characteristic humility, "I know it doesn't mean much, but I admit that these developments give me some satisfaction" (or very similar words). Adam was an academic trailblazer, and an exceptionally nice collegue, whom we all miss.

As for me, I had few teaching duties and, this time around, there was no company breathing down my neck. My work had already shifted away from formal semantics towards AI; instead of publishing in logic and linguistics, I was now much more likely to publish in Computing Science. At the same time, I found that my inclinations were closer to those of a scientist than to those of an engineer. For although I saw huge value in *implementing* one's theories – because a theory that's not implemented on a computer risks involving so much handwaving that its flaws remain invisible – it was still the theories themselves that I was most interested in. This is a fundamental difference between people: some of us are in this game because we enjoy a working piece of machinery; others are interested in grasping what makes the machine work, and what its limitations are. Both attitudes are useful, and they can complement each other well.

Although I think my time at Brighton was quite productive, I probably did not make the most of all the opportunities I was given. For instance, my colleagues and I might have engaged with teaching more. Teaching provides useful experience and is, therefore, an important item on your CV. Moreover, when ingested in the right quantities, teaching and research can reinforce each other. After all, the idea behind the modern, research-led university, which goes back to thinkers such as Wilhelm von Humboldt around 1800, is that research can inspire teaching and, conversely, teaching can inspire research.[26] To me these ideas still make sense, as does Humboldt's belief that teaching should contribute to the wider development (which he called *Bildung*, in German) of the student. And, high-minded ideas aside, some of the best memories of my working life involve moments when I felt that something I'd said or done had genuinely helped a student. Research, in my area at least, seldom gives you that.

My colleagues and I did contribute to Brighton University's teaching, but it's fair to say that the ITRI did not play a huge part in the *Bildung* of many students. I remember one day when Jon Oberlander (a well known Edinburgh-based figure in NLP who sadly died young) visited the ITRI. Having digested how we, at the ITRI, worked, he humorously concluded that we were engaged in "recreational teaching". At Brighton University, where most people outside ITRI had a daunting teaching load, our relative aloofness was arguably unwise. For a number of years, ITRI were quite successful getting research grants but, at some point, the funding tide turned, as it sometimes does. Had we engaged with the rest of the university more, then the Institute might have been rescued. Unfortunately, things turned out differently, and the axe fell in 2005.

---

[25] See https://www.sketchengine.eu for up-to-date information about the Word Sketch Engine.

[26] https://en.wikipedia.org/wiki/Humboldtian_model_of_higher_education .

## 7. Reader and Professor at the University of Aberdeen (2004-2018).

Sometime around 2003 or 2004, financial clouds started to gather over the ITRI. Judith was starting to think about moving also, so we started, very tentatively, to "dig tunnels" (as she called it). But where to go? As for me – since this workography happens to be about me – should it be a place where I could introduce a new line of work? Or would it be better to move to a department where there were many "Kees clones" already, to use another of Judith's handsomely expressive words?

I decided to accept a job offer from the University of Aberdeen, where some of my strongest peers in NLG internationally were working together, including Ehud Reiter, Chris Mellish, Graeme Ritchie, and Yaji Sripada; following a separate application procedure, Judith joined me to become a colleague once again. The people in Aberdeen did highly regarded NLG research,[27] and their interests overlapped considerably with mine. Later, I've seen people use the existence of such overlap as an argument against hiring someone. Luckily the Aberdonians took the opposite attitude, aiming to build an NLG group that had critical mass, and whose members understood each other's work.

Aberdeen's NLG group was led by Chris Mellish, a much respected figure, perhaps most famous for having written some books on PROLOG that were used widely,[28] and who now specialised in NLG. He led the group in an understated and friendly way. I vividly remember the day of my official Aberdeen job interview, during which I had to "sell" myself to people outside the Computing Science department. Before I went into the interview room, Chris and I were talking in his office. I was a bit nervous. Chris may have picked this up for, when I finally got up to leave the room he said, with a disarming smile, "The interview? You're simply going to be yourself, aren't you? I'm not worried."

At Aberdeen, I feel fortunate to have worked with Ehud Reiter, the single most influential figure in NLG worldwide who often sets the agenda, and the tone, for much of the community,[29] and whose earlier work on referring expressions was one of the reasons behind my move to Aberdeen. Once I was there, he proved to be a model of integrity and collegiality, and often the glue that held the group together. A funny moment in our interactions came early on, when I asked him for advice: Aberdeen Uni's housing department had suggested a house to Judith and me, so I asked Ehud whether he could recommend it. Did he know the area, perhaps? His answer was that he and his family had lived in precisely this house for many years. And yes, he did recommend it!

[27] A good (though slightly later) example is their work on NLG for medical decision support (Portet et al. 2009).
[28] For example, Clocksin and Mellish (1981) is a famous textbook on PROLOG.
[29] See e.g. Reiter and Dale (2000) and Reiter (2024).

Someone else I enjoyed working with very much is Graeme Ritchie, an authority in computational creativity[30] who, like Chris, had only recently joined Aberdeen from Edinburgh. Graeme's analytical mindset made him one of the best commentators I've seen; if you showed him an unfinished piece of work, he'd invariably get the point … and come up with some searching criticism, couched in innocent questions such as "Did you think of XYZ?" Putting your finger on the sore spot in a piece of work without damaging the patient is an important, and under-valued art, and Graeme had mastered this art to a T.

The move to Aberdeen, an atmospheric chunk of granite on the coast of remote North-East Scotland, worked out well for me, and for Judith as well who, at Aberdeen, grew into a substantial figure in her own area of research and teaching, and a formidable University administrator at that. Looking back, it is striking how many excellent people joined us there over the years, as lecturers, postdocs, or as PhD students. For me, the move was also my first acquaintance with a normal academic teaching load, focussing on areas of Discrete Mathematics that are relevant to computer science, such as the theory of computability. What worked particularly well, in the classes that I taught, is combining the challenging concepts of computability with some simple programming tasks (using the functional programming language Haskell, because of its ability to perform calculations on infinite sequences), which helped students to gain a hands-on understanding of these concepts. Even the marking was somewhat interesting, because students regularly found clever new solutions to problems.

Teaching is serious business: for better or worse, you're playing a key role in a young person's life. (And if you forget to turn up one day, then a classroom full of people are going to notice it.) Teaching becomes even more serious when the local culture exalts administrative process. In Aberdeen, as in most UK universities, exam questions, for instance, had to be vetted by both an internal commentator and an external examiner several months before the exam. Consequently, the lecturer needed to have their exam questions ready at a point in time when their course has only just started. This can be difficult, of course, particularly if you're teaching the subject for the first time.

For me at least, a baffling aspect of academic life in the UK was the grading procedures that are employed for judging students' exams and coursework, which are best thought of as akin to the old Imperial System of weights and measures. In The Netherlands, teachers at all levels grade a students' work with a number between 1 and 10, using 5.5 as the lowest pass mark. This may not be ideal, but it has the advantages of clarity and stability. In the UK, by contrast, a variety of schemes are in place, so a new teacher or lecturer has to get used to whatever scheme a particular school or university happens to use. Many institutions insist on using a plurality of schemes, some of which are more fine-grained than others. Let me explain. (And if it sounds convoluted, then that's because it is.)

[30] An influential monograph in this area is Ritchie (2004).

When I started at Aberdeen, staff were marking each exam paper out of (in the simplest case) 100; the resulting grades were translated into a supposedly more didactically meaningful, non-linear, 20-point scale. The grand total of a student's grades was then finally translated into a final degree classification, yielding such verdicts as "First Class", "Upper Second Class", "Lower Second Class", and "Third Class", which would end up on students' official graduation record, and which are important to employers. (Why only the Second Class is split into two parts is a question for historians.)

To make things worse, the entire grading procedure is subject to change. In my 14 years at Aberdeen, I've seen several such changes. In 2014, for example, the 20-point scale was changed into a 22-point scale, later to be extended to 23 points; in further changes, the crucial pass-fail boundary fluctuated between 40, 45 and 50 out of 100, depending on the policy of the day. Since grades started out on the most fine-grained scale, then had to be "translated" into successively more coarse-grained scales, we as educators were confronted with two sets of decisions that could be hugely consequential for the students, and which were therefore taken very seriously. Confronted with a particular final-year exam paper, my reasoning might go, "So-and-So is scoring 59 out of 100 for this exam, yielding a grade of Lower Second Class; if I boost her mark to 60, that will yield an Upper Second. Does she deserve this, or should I stick with my original assessment?" This was not some convoluted thinking solely on my part: we were encouraged to always "check borderlines" between degree classes because much depended on it for the student.

I felt ambivalent about this teaching culture. On the one hand, I admired my colleagues' seriousness, and their very genuine engagement with students' welfare; and, changes in the marking procedure were always made with the best intentions. On the other hand, the cost in terms of academics' time, and hence taxpayers' money, were substantial. On a bad day, the whole thing felt to me like an expensive game, which was played just for the sake of it. I came to grudgingly accept it as one of the necessities of life, until I experienced a more laidback teaching culture in Utrecht (Section 9); I could be wrong, but I don't think the students at Utrecht are shortchanged as a result of Utrecht's substantially less onerous procedures around exam setting and marking.

Outsiders tend to equate being a lecturer with explaining potentially difficult subject matter, but I discovered that the effectiveness of your teaching depends more on your being both enthusiastic and, especially, well-organised than on your ability to explain difficult things. Because, let's face it, most of the things you'll be explaining have been explained much better, by others. Their work is often publicly accessible: in books, on web sites, and in educational video clips on YouTube and elsewhere. So, if you're a starting lecturer and you feel insecure about your lecturing – for example because your mastery of the topic is still a bit feeble, or you're lecturing in a language that isn't yours – then don't despair: as long as you offer students a clear time schedule, and you make all past exam papers and other course materials available to them, they'll erect a statue for you, even if they don't understand a word you say about the academic subject matter of the course. (I'm probably exaggerating, but you catch my drift.)

My research continued to focus on "AI as a science", rather than "AI as engineering" (see Section 2). Increasingly, I learnt to combine theories, algorithms, and experiments, so as to build empirically supported computer models of human language use. The ambition behind this approach is to build models that are more detailed, and more "testable", than the theories of language production developed by most linguists and psychologists.

At some point, Ehud brought me into contact with Roger van Gompel in nearby Dundee, whose expertise in psycholinguistics ended up strengthening this line of work considerably, by enhancing its experimental and statistical rigour.[31] Together with Albert Gatt, who had joined me as my PhD student from Brighton to Aberdeen, and with Emiel Krahmer, who was and is running a research group at Tilburg University in which NLG takes a central place, we embarked on a lengthy investigation of human reference production. It became one of the most rewarding strands of work I've been involved in. Each of the four of us contributed in their own way, allowing everyone to benefit from each other's strengths.[32] Further crucial contributions were made by PhD students and postdocs including Ivandré Paraboni, Imtiaz Hussain Khan, Roman Kutlak, Margaret Mitchell, Ielka van der Sluis, and Xiao Li.

One thing that helped to make this work successful is how, time and again, Emiel and others managed to find opportunities for organising research workshops on the topic of referring expressions, which gave us opportunities to meet up with each other, and to get to know other researchers interested in our work. This was all the more crucial because it allowed us to interact with people who approached our topic in different (e.g., non-computational, purely experimental) ways, building bridges between disciplines.

The main upshot of this line of work was broadly two-pronged. First, we demonstrated how to build algorithms that refer effectively in complex and challenging situations.[33] Second, we brought a new level of rigour to NLG algorithms, combining established psycholinguistic methods with a centralised approach to evaluation that had become commonplace in some areas of NLP but that had not been applied to NLG algorithms yet. The idea of the latter is to not rely solely on research groups individually evaluating their own algorithms, but to organise a centralised evaluation "campaign" in which research groups from around the world are invited to submit their algorithms, all of which are then tested systematically, following a publicly available protocol. After the first campaign, in which Albert Gatt was the main driving force on our side, evaluation campaigns became a fixture at the yearly NLG conferences.[34]

---

[31] Later on, Ellen Bard, a psycholinguist at Edinburgh University who specialized in the study of reference among many other topics, helped to reinforce the experimental aspect of our work even further.
[32] See Van Gompel et al. (2019), for example, where our algorithm called Probabilistic Referential Overspecification (PRO) is motivated and evaluated.
[33] Reference is challenging, for example, when the target referent is a set of entities (rather than one single entity), or when the speaker and hearer have divergent information about the domain, or when the domain is so large that finding the referent becomes challenging for the hearer.

Successful research often starts from simplifying assumptions. Accordingly, our work had always looked at communication in relatively simple situations. For example, where our experiments involved objects of different sizes, we usually made sure that the objects came in two sizes only, the large ones and the small ones, where all the small ones were of the same size, and all the large ones too; similarly, all the red objects were the exact same shade of red; and so on. This strategic decision allowed us to ignore the challenges that arise when objects come in a continuity of sizes, colours, and so on.

In order to understand what happens in more realistic situations, I started a separate line of research into logical and computational models of vagueness – a longstanding research issue in formal semantics[35] – building on some preliminary work that I'd done while at Stanford (Section 5). A word or phrase is called *vague* if it is defined in such a way that borderline cases can arise. In ordinary language, this happens often. For instance, if I tell you that John and Mary arrived in the bar "at the same time", you're unlikely to understand me as asserting that the two arrived at exactly the same time; I'm claiming that their arrival times were close together; *how* close together is not cast in stone. Consequently, borderline cases can arise (e.g., if John arrived at the bar just half minute after Mary) for which it is undefined whether my claim was true or false.

I felt that these issues are so pervasive, and some of the theories so elegant, that it would be worthwhile trying to write a book about them that appeals to specialists and non-specialists alike. The book that I ended up writing[36] discusses a variety of techniques designed to do justice to the vagueness of expressions, defending a probabilistic approach. Pitching these ideas in down-to-earth language was hard work, but the effort paid off; the book sold well and I came to feel there might be issues that are best addressed in this "pop-sci" format, which strips away jargon and reduces issues to their core.[37] Not every issue is easily handled in this way, of course. For example, the book did not explain why vagueness is so pervasive in human languages.[38] Our own later experiments with human participants suggest that this is probably not because vagueness adds some kind of utility for hearers, but only for other reasons such as, for example, the discriminative limitations of our eyes and ears.[39]

---

[34] See Belz et al. (2010) for discussion of the evaluation campaigns for referring expressions generation.
[35] See e.g., Sorensen (2023) for background.
[36] Van Deemter (2010).
[37] I think the discussion of *epistemicist* ("vagueness as ignorance") solutions to the Sorites Paradox is a good example of what an informal treatment can contribute (Van Deemter 2010, Chapter 7).
[38] See also Lipman (2000).
[39] Green and Van Deemter (2019).

I've never been very interested in academic promotions. Why apply to become a full professor when you're happy in your work, and free to focus on any research topic you like? I was, in other words, perfectly content with my position as the ant in the Daoist philosopher's story (Section 5). But one episode, around 2010, made me change my mind. I'd seldom quarrelled with colleagues, but suddenly one of my administrative roles – I was responsible for monitoring the progress of the PhD students in the department – brought me into conflict with a powerful figure in the department who, being a full professor, outranked me (since I was still a Reader). In a nutshell, it had become apparent that some of his PhD students had not made much progress, and my monitoring threatened to put a spotlight on this situation. Unfortunately, my colleague decided to defend himself by turning on me. It was an unpleasant episode, and although I think it's fair to say I won this battle, if I had lost then this would have put me in a vulnerable position. At last, I realized that "being an ant" is not always a good thing, so I decided to apply for a full professorship. Given that it was relatively late in my career, I had reason to expect – correctly, as it happened – that my application would be successful.

For the largest part of my career, I kept managerial responsabilities to a minimum. Yet, over time, my duties did shift subtly towards the "admin" side of things, as is true for most academics. One aspect of this was hiring personnel, a task that is decisive for the future of a department, and very challenging at the same time. How do you judge a person when all you have to go by is some paperwork, a research talk, and perhaps a short visit? How do you do this fairly and without expending huge amounts of time?

Because hiring is time intensive, committees often make use of bibliometric services such as Google Scholar, which lists a researcher's publications, including citation counts for example. Scholar is a wonderful device if it's employed wisely. To see how it can be useful, look up the Scholar page[40] of the famous Alan Turing and click on his 1950 paper "Computing Machinery and Intelligence", which I mentioned in Section 1. The page shows clearly that, far from being forgotten after all these years, (as will be the case with everything that we, mere mortals, write) its year-on-year citation counts still grow rapidly.

When preparing for an interview, I look at what a candidate has published. Are their yearly overall citation counts going up or down? How often are they (still) the first author on their publications? And so on. Now, when you meet the candidate, you can ask focussed questions such as "Do I understand correctly that the focus of your research has shifted from logic to NLP recently?", or "You've been working with the same set of co-authors for a long time now. Could you tell us a bit about everyone's role?" Or, "We're fascinated to see your work has been taken up by researchers in medical research. What is it that they're finding useful in your work?" In this way, you can have a more meaningful discussion than would otherwise be possible.

[40] https://scholar.google.com/citations?user=VWCHlwkAAAAJ&hl=en&oi=ao The paper was reprinted in 1995.

On the other hand, bibliographic metrics should be taken with a pinch of salt. One reason is that researchers may be tempted to "play games". One popular game is what I call *the generosity game for authors*. Here's how it works: When you publish a paper, you ask others to become co-authors. Not just a few people; you should be generous, inviting everyone who has had even the slightest input into the research. Your generosity costs you nothing, because all the main metrics ignore the number of authors (and the order in which they appear). On the other hand, your generosity has no end of benefits. Your co-authors will put the paper in their institutional repository or on their web page; they may even mention the paper on social media. All of this is a free advert for your work, leading to more citations for you. Your co-authors are also likely to refer to the paper in their later work, and this will boost their own citation numbers too (because most metrics do not penalize self-citation). Best of all, if they're really talented players of the generosity game, they'll reciprocate by inviting you to become a co-author on some of their later papers, and this will boost your citation numbers even more. Everyone wins! Note that the generosity game does not require any sinister "pacts" between authors to make you and your friends look better than you are; human nature does most of the work.

Another weighty academic responsibility is judging proposals for research project funding, for example by national research councils or the EU. Sometimes when I'm sitting on a panel to judge proposals, I initially feel some trepidation, because my expertise is somewhat lacking (e.g., because the proposal makes use of some theories or techniques that I'm relatively unfamiliar with); yet, when I'm on my way back home, I often realize that most people in the meeting understood less of the proposal than I. Typically, it's not that my fellow panel members are lacking in experience, but their understanding of the subject is less impressive sometimes.

Of course it's a good idea to have some experienced people in the room, because they bring the kind of wisdom that builds up over time. Their wisdom, however, should always be complemented by area-specific knowledge and by the kind of technical skills that tend to deteriorate, not grow, with age. Yet funding and recruitment panels are often packed with very senior people, whereas experts – as in: people who (still) do research in the same area as the applicant – are deployed much more sparsely. As a result, important decisions are sometimes based on highly salaried gut feelings more than on genuine expertise. I don't see how this can be the best way to go about it.

Good leadership implies respect for differences. I do not only mean cultural or racial differences or differences in lifestyle, where mutual respect is obviously crucial. What I mean also is differences between academic disciplines, and even within a given discipline. Scientific work is often characterised by a plurality of approaches and methods. This is certainly true in AI, where there exist huge differences in attitudes and values between, for example, the more application oriented and the more theoretically inclined; between algorithm builders and those who test and evaluate algorithms; between people who work on different AI tasks; and so on. Therefore, it is important to be able to value work that's very different from your own. If you're not, you'll suck at hiring people, at judging funding proposals, journal and conference submissions, and at setting up research collaborations with people outside your own narrow circle.

## 8. Visitor in China (Harbin, Beijing, and elsewhere)

For all its drawbacks – because travel is time and fuel intensive – we sometimes need travel to escape our own bubble. Some of my most rewarding international collaborations have been with people in China, starting around 2007, when I was based in Aberdeen.

Our collaboration started with Professor Tiejun Zhao, of the famous Harbin Institute of Technology (HIT), approaching our Natural Language Generation (NLG) group in Aberdeen. Having introduced HIT as one of China's foremost players in NLP, Tiejun declared that HIT was relatively weak in NLG. Since our group was strong in that particular area, he was proposing for the two groups to collaborate, starting with a series of summer schools in NLP, in which Microsoft Asia, based in Beijing, took part as well. Research workshops with partners in Hangzhou, Guangzhou, and Xining were to follow. Since you're a serious reader, I'm not going to bore you with trivia, such as my nocturnal fall into Hangzhou's celebrated West Lake. (Just ask me the next time you see me!) One work-related anecdote won't hurt though…

Once, after lunch, Tiejun Zhao was addressing a large lecture theatre full of students who had travelled to Harbin from all over China. I'd had a rough time that morning, battling jet lag and struggling to find the lecture hall on a campus as large as a medium-sized town, so when I finally arrived in the hall, I sank into my seat. Safe in the knowledge that I had no teaching duties that day, I may have dozed off for a second. At some point I heard Tiejun say, loudly and clearly – perhaps because he felt he had to repeat himself – "Now it's Professor van Deemter's turn. For the remainder of this afternoon, he'll lead a workshop on paper reviewing". I got the shock of my life, because I had not prepared anything! I remember ambling to the front of the room, scrambling to think up a suitable plan B. After ten confusing minutes, we found a solution that worked out reasonably well, or so I'd like to believe. That evening in a restaurant, Tiejun and I tried to figure out how this misunderstanding between us had arisen. I don't think we ever found the answer, but in the end we laughed it off.

The most lasting inheritance from these collaborations resided not in those mythical sums of Chinese government funding that some people dream of, but somewhere else entirely. Through my travels, I befriended people and started learning some Mandarin. Gradually, I gained a professional interest in the language as well, and I started investigating the differences between languages, focussing on cases in which speakers in one language express information that speakers of the other language *omit* (e.g., when English speakers mark a noun phrase as plural whereas Chinese speakers leave its number unspecified). This research is still ongoing, together with my former PhD student Guanyi Chen, who works in Wuhan,[41] and Rint Sybesma, a sinologist at Leiden University. Our work does not fit the current mainstream of application-oriented AI (see Section 11), but this is a price we are willing to pay in exchange for the intellectual rewards we're hoping to gain.

One of the lessons that I've learnt from my work-related travel is that researchers across the globe are incredibly similar in their knowledge, skills, manner of working, and general outlook. Suppose you board a plane on your way to a conference and you land in a foreign country where, at first sight, everything appears to be different. Then, once you've managed to find the venue, you enter a room, and you start interacting with the people in it. Suddenly everything is familiar. Because, regardless of where they are from, the people in the room have read the same research papers as you, and they've had a very similar education. It's almost – well, almost – as if you'd never left your home.

Sometimes, the feeling of being at home in a faraway country can be particularly strong. On a cold and dreary Friday afternoon in March 2017, I arrived in my flat in Beijing to start my sabbatical at the Chinese Academy of Sciences, where I was going to join the group of Professor Le Sun. My flat was in a sprawling and unfamiliar neighbourhood in the north of the city, full of grey highrise buildings. After a somewhat disorienting walk around my new neighbourhood, I sent email to two old acquaintances from Aberdeen, Guanlin ("Epsilon") Li and Xiantang Sun, who had returned to China after spending time in Scotland, and whom I had not been in touch with for some considerable time. I told them about my new environment, hoping to get some sympathy from them. Within a few hours I got their responses, and what they told me was nothing less than astonishing: both people happened to either live or work in Beijing, and within a range of 500 yards from my flat! Both of them were happy to meet me. Within half a day, my lonely existence turned into a thoroughly enjoyable episode filled with joint meals and outings. I still marvel at my luck: in a country as huge as China (and even in densely populated, and research-obsessed, northern Beijing), what are the chances?

Sometimes, of course, one does encounter some genuine differences between countries. For example, my connections with people in Harbin and Beijing gave me a fascinating window onto a country that had understood early on that the future of AI was going to be based, at least for some time, on purely statistical techniques. The use of statistical techniques was very common in the USA and Europe too, and this included our own work in Aberdeen as well, but the single-mindedness of my Chinese colleagues' focus on such techniques, avoiding the use of rule-based methods or linguistic insights of any kind, came as a surprise for me nonetheless.

Collaborations with China are currently under pressure for political and other reasons, yet I really hope that others, in both East and West, will get similar opportunities for broadening their outlook to the ones I had.

[41] Guanyi Chen's PhD thesis is still a good introduction to this type of work (Chen 2022).

## 9. Professor and Emeritus at Utrecht (from 2018)

One day in 2017, during my sabbatical in Beijing, I got a message from Judith: How would I feel about moving back to The Netherlands, more precisely to Utrecht? There was an interesting job opportunity for her there, and she wondered whether or not to explore it. I responded without giving the matter much thought, writing something like, “Why not? Utrecht is pretty and we have friends there.” “Sure”, she wrote, “I’ll keep you in the loop. Brexit is coming, so maybe it’s time to go home?”

Utrecht’s Computing Science department had an impressive track record in theoretical computing science[42] and were now diversifying into other areas. Judith would be well placed to help this process along, for example by extending and solidifying empirical aspects of Computing Science, such as the experimental underpinning of AI algorithms and systems. Having met all the relevant people in the department during some short visits, she decided she wanted the job.

Following my return from China, I visited Utrecht three times, after which the then Head of Department, Johan Jeuring, offered me a professorship as well, rather gallantly leaving it to me whether I wanted to start an NLP group or not. There were no local “Kees clones” this time, but there were pockets of NLP research in Computing Science, and others at Utrecht’s excellent Language Sciences department, and elsewhere. I could live with this situation, so now Utrecht looked like a realistic option for both Judith and me.

Academics often lead nomadic lives, and this makes the two-body problem, as it is sometimes called, quite common. I felt at home in Aberdeen, where Ehud Reiter and I had been taking turns leading a vibrant NLG research group. How would things be in The Netherlands? Unlike Scottish universities, Dutch ones use a mandatory retirement age, so my job at Utrecht wouldn’t last for more than 6 or 7 years. Moving back to The Netherlands, in other words, would mean something quite different for me than for Judith. However, in the past, Judith had followed me to Brighton and Aberdeen, so who was I to throw a spanner in the wheel now that it was she who wanted to move?

We decided to take the plunge, and we moved to Utrecht in the Summer of 2018. I was unsure about the idea of starting a new group and I left the matter undecided for a while. However, my PhD students Guanyi Chen and Linda Li came with me. Among other things, we turned to some issues that had occupied me at the start of my career. Back in the 1980s, I had used logic to formalise the meaning of sentences in ordinary language; now we approached the matter from a computational angle: we investigated the ability of NLG systems to use quantifiers (like “all”, “most”, “five or six”, “equally many A as B”) to best effect in generated text. An interesting aspect of this work, further addressed in the PhD project of Eduardo Calò, is how ordinary languages are surprisingly effcient in their expression of logical information, for example by the clever use of words like "each other", "respectively", "only", and "exactly". Studying these logic-related issues made me feel a bit as if my career had come full circle.

[42] See e.g. Van Leeuwen (1991), Bodlaender (1998).

In other respects as well, the move to Utrecht proved fortuitous. Slowly, flying under the radar, AI was becoming fashionable once again, and student numbers were rising, in Utrecht as much as anywhere else. The year was 2018, about 5 years before the introduction of prompt-based NLP systems such as ChatGPT, yet neural "Deep Learning" methods – which took their inspiration from neural mechanisms in the brain – had become astonishly successful, even on tasks that were very different from the ones the system had originally been trained on.[43] New government initiatives were launched for boosting research in AI and other areas of computing. Soon there was money to go around. Between 2020 and 2024, the Computing department embarked on a period of spectacular growth: in one year alone, as many as six new full professors were hired, all of whom were working in AI or in areas closely associated with AI.

Consequently, the atmosphere turned out very different from earlier years, when AI research was often starved of money (Sections 4 and 6 above). In earlier jobs, I'd become used to seeing the ceiling come down, but at Utrecht, nothing of the sort happened. From 2020 till 2023, the world-wide Covid epidemic made life difficult of course, particularly for those of us with elderly relatives or family abroad, but other than that, the main challenge facing us was: coping with the explosive growth of AI.

Initially, in Utrecht, much of the new AI money bypassed NLP. This was partly because of my own ambivalence, in those early days, about whether or not to grow a new NLP group, and partly because some of my new colleagues took every opportunity to explain that NLP was "not really" AI – that old chestnut – and hence not as deserving of investments in AI as their own work.

But ultimately, the new wave of neural NLP proved hard to ignore, for them and for me, the more so because many students insisted on learning about this newly popular area of research. Soon, in a funny reversal of history (see Section 6 about Doug Lenat's views), our classrooms came to be populated by youngsters who considered neural, Deep Learning-style NLP to be the only "real" AI. At the same time, our departmental NLP group grew rapidly. Dong Nguyen, whose work in computational social science[44] was widely acclaimed already, joined the group, followed by Gerard Vreeswijk, Marijn Schraagen, and Pablo Mosteiro among others, and a growing number of PhD students, including Fahime Same, who worked with me from Cologne. In a further big boost, Massimo Poesio joined us from London in 2023 to lead a project in which 5 PhD students study meaning variation in NLP. We collaborate with many other people in Utrecht, both within and outside the Computing Science department, perhaps most intensively with

[43] See e.g. Devlin et al. (2019) on the BERT system.
[44] See e.g. Nguyen et al. (2016, 2020).

Rick Nouwen,[45] Denis Paperno,[46] Antal van den Bosch, and others in the Language Sciences department, whose interests fit our own like a glove. Now, after my retirement, our group is led by Albert Gatt, my former PhD student, who is more suited to this role than I ever was, and whose arrival from Malta in 2021 added a multimodal aspect to our research, to do with the relationship between images and text.[47]

Back in 2017, when I was pondering our move to Utrecht, it was impossible to foresee that a new, successful NLP environment would emerge here so organically.

How long will the good times last? If eight years has been a long time in academia, it's been an *aeon* in politics. Suddenly, the wider environment in which we work contains significant threats. The war in Ukraine demands huge resources. At the same time a new populism is at large, which opposes immigration, and which looks upon overseas students as if they were a burden rather than a blessing. Respect, and funding, for research and education are under threat. Similar conditions exist across much of the globe,[48] but the risks of intellectual and economic decline seem especially big in a small country like The Netherlands which, at 18 million inhabitants, is less populous than quite a few cities, and which is dependent on international collaboration in almost every aspect of life.

Fortunately, these risks have not fully materialised here yet, especially in areas such as AI, which happen to be in vogue. (Academics in the humanities are less lucky!) Utrecht's NLP group, for example, is still growing and outward looking, counting 11 nationalities among the 19 people listed on its web page. This is not by design of course, but in the same way that football teams have become multinational, trying to hire the best talent their money can buy. In a small country, when hiring is fair and skills based, and when the skills involved are not tied to a particular language or national culture, then internationalisation is the unavoidable – and, incidentally, quite enjoyable – outcome. In such an environment, the only alternative to globalisation is stagnation.

Despite the threats that are facing us, I can still wholeheartedly recommend a career in academia. It has been said that, in other walks of life, everything is either politics or identity. Perhaps I'm naïve – I see some readers nodding – but in my limited experience, science is different. Few people are deeply interested in your politics, your gender and sexual orientation, your religion, or the tone of your skin; what they care about is your work, and whether you do it in an honest and collegial manner. (And, obviously, whether you're available for a game of table tennis in the canteen.) This single-mindedness is a trait that I've come to appreciate more and more. I used to ridicule the Harry Potter-like formalities that are so prominent in Dutch academic life, where professors wear silly robes and hats even at such semi-official events as a PhD defence; but lately, now that so much public discourse is driven by feelings more than facts and reason, I've come to almost *like* these paraphernalia of academic authority. Perhaps at a time when all expertise has become suspect, our robes and hats can remind us of our calling.

[45] See e.g. Wong et al. (2025).
[46] See e.g. Paperno et al. (2014).
[47] E.g., Parabelescu et al. (2022).
[48] For example, a recent editorial in *Nature* denounced what it calls an "unprecedented attack" on research and research infrastructure in the USA during the first month of the second Trump administration (Editor-Nature 2025).

The Dutch retirement age currently stands at 67. Although Utrecht University does not encourage their employees working beyond that age, the recent award of a sizable new research project made it tempting for me to go on for a bit, albeit in a reduced capacity. It seems to me, however, that in a fast-moving field like AI, retirement makes sense. Looking around, it appears to me that my colleagues' grasp of, and interest in, the most recent types of AI tends to decrease with age. Life experience, by contrast, increases with age. Perhaps it's time for me to trade work for life.

## 10.Perverse Incentives

A friend of mine, who works in a different academic discipline, read a draft of this workography and observed how fortunate I had usually been with my colleagues. His implied question was: How can anyone be so lucky? On reflection, I have to admit that the academic coin does have a darker side.

One day, somewhere, my colleagues and I were walking towards a canteen to grab some lunch. We were discussing a certain Mr X, who was known to put his name on research articles to which he had not contributed; the Head of Department seemed to be turning a blind eye, perhaps because X was, apart from everything else, quite smart and productive. Now, apparently, X had leaned on one of his postdocs to make it look as if he, X, had been a major contributor to a huge funding proposal that it had taken other people ages to compose. After an embarrassed silence, I said wrily, "You wonder whether there's anything that X would not be prepared to do to boost his CV. Under the right circumstances, would he be prepared to kill?" Silence again. After a few seconds, the colleague walking next to me looked at me and said, "We're not laughing, are we?" I had meant to exaggerate, but for a brief moment it seemed that an affirmative answer to my question was not beyond the realm of possibility.

In my experience, the overwhelming majority of academics are scrupulously honest, so our lunchtime conversation was, in large part, about Mr X's unique personality. On the other hand, Mr X is not alone. We, academics, can be vain, and driven by so many "perverse incentives" – to do with publishing papers, being cited, obtaining research funding, heading research groups, joining lofty Societies, and so on – that hyper-ambitious behaviour is not exactly rare, and accidents do happen (although luckily, in most cases, the victims live to tell the tale).

Perverse incentives mean that there is no lack of *motive* for academic transgressions. Moreover, senior academics can hold considerable power over younger colleagues, so they often have *opportunity* as well. Universities have become increasingly aware of these pitfalls, particularly after some well publicized “me too” affairs. The measures that have consequently been taken in many places – for example, ensuring that every PhD student has two supervisors rather than just one – should help to reduce the problem.

## 11. Wrapping up: Artificial Intelligence and NLP in 2025

In this workography, I have sketched my 40-year path through an exciting and rapidly changing area of research and teaching. I started out my working life as an AI engineer until, some time around the year 2000, I re-discovered the more theoretical interests that had drawn me towards formal semantics and AI as a student (Section 3). Over roughly that same period, AI went from a mirage somewhere at the edges of Computing Science to the worldwide phenomenon that it now is. The “AI winter” (Section 2) is thus well and truly over; AI is thriving more than ever before.

In the age of Deep Learning, AI has made huge strides solving many practical tasks, ranging from data analysis and decision making to image processing and a wide range of NLP tasks, such as Question-Answering, Machine Translation, and automatic text summarisation, for example. Yet, on the other hand, these practical successes carry huge risks, the more so when AI – including bots that generate disinformation, for example – ends up in the hands of immoral agents. These are risks I had vastly underestimated when I entered the field. At the time of writing, they’re hard to overlook.[49]

I have a pet intellectual worry as well: Today’s AI does not yet excel at the things I’m most interested in. I’m disappointed, for example, that the ethos of AI research is now so predominantly an engineering ethos, which ignores the contributions that computational models can make to our understanding of human behaviour.

Current AI models in the tradition of Deep Learning, such as ChatGPT for example, are hugely impressive “machines” but, by themselves, they do not tell us very much about human language and communication that, for example, a linguist or a psychologist would be interested in knowing. This is partly because the bulk of today’s NLP tells us little that can claim to offer good *explanations* for the intricacies of human language use.[50] Luckily, these are issues that some of my colleagues are trying very hard to address, for example by working on techniques that make it possible to gauge what a given model has actually learnt and why the model takes certain decisions; by, either mathematically or experimentally, investigating the limitations of a given type of model;[51] by working on hybrid systems that combine Deep Learning with rule-based approaches; and by using a Large Language Model not as an end in itself but as a tool for addressing scientific questions about language.[52] A related type of work investigates the extent to which Large Language Models can be seen as models of human cognition.[53]

[49] See e.g., Van Miltenburg (2025) for a report undertaken for the ACL’s Special Interest Group on NLG (SIGGEN).
[50] See Van Deemter (2023), which discusses the explanatory value of NLP models.
[51] Good examples of this type of work include Merrill (2021), Xu et al. (2024).

It is impossible to say how successful each of these new endeavours will be in the longer run, but I hope that AI can shake off its remaining limitations as convincingly as it did the ones that caused the "AI winter" of the last century. If this can happen in such a way that scientific aspects of AI take pride of place once again, with due consideration for the many ways in which AI can shed light on human language and communication then, for me at least, that will be a consummation devoutly to be wished.

**Acknowledgement**: I'm grateful to colleagues who gave me feedback on a draft of this paper. One influential comment came from Eduardo Calò, who said, "More of those anecdotes, please!" So, here it is … with even more anecdotes.

## Reading Matter

[52] See Potts (2023) for example, which uses LLMs to study a famous problem in Theoretical Linguistics known as "the poverty of the stimulus". See also Kallini (2024) for another interesting strand of LLM-based scientific work.
[53] See e.g., Goldstein et al. (2024).

---

Kees van Deemter,
Utrecht University, 5 April 2025; slightly revised on 29 July 2025.
Comments to c.j.vandeemter@uu.nl please.